\documentclass{article}
\usepackage{microtype}
\usepackage{graphicx}
\usepackage{subfigure}
\usepackage{tcolorbox}
\usepackage{booktabs} 
\usepackage{xcolor}  
\usepackage{microtype}
\usepackage{graphicx}
\usepackage{subfigure}
\usepackage{algorithm}
\usepackage{algpseudocode} 
\usepackage{longtable}
\usepackage{algpseudocode}
\usepackage{multirow}
\usepackage{booktabs} 
\usepackage[font=small,labelfont=bf]{caption}
\usepackage{hyperref}

\usepackage{amsmath}
\usepackage{amssymb}
\usepackage{mathtools}
\usepackage{amsthm}

 \usepackage[preprint]{neurips_2025}

\usepackage[utf8]{inputenc} 
\usepackage[T1]{fontenc}    
\usepackage{hyperref}       
\usepackage{url}            
\usepackage{booktabs}       
\usepackage{amsfonts}       
\usepackage{nicefrac}       
\usepackage{microtype}      
\usepackage{xcolor}         

\title{DeepACTIF: Efficient Feature Attribution via Activation Traces in Neural Sequence Models}

\author{%
  Benedikt W. Hosp \\
  University of Tuebingen \\
  hospbene@gmail.com \\
}

\begin{document}

\maketitle

\begin{abstract}
Feature attribution is essential for interpreting deep learning models, particularly in time-series domains such as healthcare, biometrics, and human–AI interaction. However, common attribution methods like Integrated Gradients or SHAP are computationally intensive and not suited for real-time applications. We present DeepACTIF, a lightweight and architecture-aware feature attribution method that leverages internal activations of sequence models to estimate feature importance efficiently. Focusing on LSTM-based networks, we introduce an inverse-weighted aggregation scheme that emphasises stability and magnitude of activations across time steps. Our evaluation across three biometric gaze datasets shows that DeepACTIF not only preserves predictive performance under severe feature reduction (top-10\% of features), but also significantly outperforms established methods including SHAP, IG, and DeepLIFT in terms of both accuracy and statistical robustness. Using Wilcoxon signed-rank tests and effect size analysis, we demonstrate that DeepACTIF yields more informative feature rankings with significantly lower error across all top-k conditions (10–40\%). Our experiments demonstrate that DeepACTIF not only reduces computation time and memory usage by orders of magnitude but also preserves model accuracy when using only top-ranked features. That makes DeepACTIF a viable solution for real-time interpretability on edge devices such as mobile XR headsets or embedded health monitors.

\end{abstract}

\section{Introduction}
\label{sec:intro}

As machine learning models become increasingly integrated into decision-making processes, the demand for interpretable and efficient AI is growing. In domains like healthcare, biometrics, and human-computer interaction, the ability to understand model predictions is critical for building trust and ensuring reliable outcomes. Feature importance analysis, which quantifies the contribution of individual input features to model predictions, plays a key role in achieving interpretability. However, traditional methods such as Integrated Gradients \cite{sundararajan2017axiomatic} and Shapley values \cite{lundberg2017unified} are computationally demanding, making them impractical for real-time applications or deployment in resource-constrained environments.

Long Short-Term Memory (LSTM) networks are widely used for sequence modelling due to their ability to capture temporal dependencies. Yet, their sequential and high-dimensional nature complicates the estimation of feature importance, mainly when interpretability must be achieved under strict efficiency constraints. These challenges are further magnified in real-world settings such as medical diagnostics, where decisions must be interpretable and delivered in real-time on low-power devices.

This problem is also relevant from a sustainability perspective: the computational cost of explainability methods contributes to the growing environmental impact of AI systems. Efficient attribution methods are thus crucial not only for practical deployment but also for reducing energy consumption and aligning with principles of sustainable AI.

In this work, we introduce \textbf{DeepACTIF}, a novel framework for feature attribution in LSTM-based sequence models that combines interpretability with computational efficiency. DeepACTIF leverages internal activations to derive feature importance scores, eliminating the need for gradients or perturbations. Its core innovation is an \textit{inverse-weighted aggregation strategy}, which prioritises features that exhibit consistently strong activations across timesteps and samples, leading to stable and robust feature rankings.

We empirically evaluate DeepACTIF across three biometric gaze datasets, comparing it to state-of-the-art methods including SHAP, IG, DeepLIFT, and ablation-based approaches. Our results show that DeepACTIF significantly outperforms these baselines in preserving model performance when reducing to the top-k \% most important features (for $k \in \{10,20,30,40\}$). Statistical tests confirm that these improvements are robust, with Wilcoxon signed-rank $p$-values consistently below 0.05 and large effect sizes (Cohen’s $d$), validating DeepACTIF’s reliability and utility.

The remainder of this paper is structured as follows: Section~\ref{sec:related_work} reviews related work in feature attribution. Section~\ref{sec:method} details the DeepACTIF method and inverse weighting strategy. Section~\ref{sec:experiments} describes our experimental setup. Section~\ref{sec:results} presents quantitative results across datasets and methods, and Section~\ref{sec:discussion} discusses implications for scalable, interpretable AI. We conclude in Section~\ref{sec:conclusion}.

\section{Related Work}
\label{sec:related_work}

Interpreting deep neural networks remains a critical challenge, particularly in time-series and sequential domains where both temporal dependencies and high input dimensionality complicate attribution. Feature attribution methods can be broadly categorised into three classes: gradient-based, perturbation-based, and activation-based or aggregation strategies.

\textbf{Gradient-based methods} such as Integrated Gradients (IG) \cite{sundararajan2017axiomatic} and DeepLIFT \cite{shrikumar2017learning} compute feature attributions by analysing the gradients or differences in model outputs with respect to inputs. IG performs integration along a path from a baseline to the input, while DeepLIFT tracks activation differences relative to a reference. These methods are popular due to their theoretical appeal. Still, in practice, they are sensitive to baseline choice and suffer from high computational cost, especially when applied to recurrent models like LSTMs. Their runtime complexity makes them unsuitable for time-critical or embedded settings.

\textbf{Perturbation-based methods}, such as Ablation \cite{Meyes2019Ablation} or Permutation Importance \cite{Radivojac2004Feature}, assess a feature’s relevance by observing the model's response to input modifications. Although conceptually intuitive, these approaches require repeated forward passes, scaling poorly with feature count and sequence length. Moreover, in temporal models, they risk breaking sequence structure, resulting in less reliable or misleading attributions \cite{Tjoa2019A, Shawi2020Interpretability}.

Aggregation strategies attempt to combine per-sample attributions into a global ranking. Many existing works rely on averaging across timesteps or input samples, implicitly assuming stable contributions over time. However, this can fail in sequence models, where feature relevance is context- and time-dependent. Simple averaging thus risks masking important dynamic patterns and overrepresenting noisy activations.

Recent evaluations \cite{SurveyTimeSeries2023, EvaluationTimeSeries2023} underscore these limitations, particularly in LSTM-based models used in biometric or health-related tasks. Comparisons between SHAP and IG \cite{ShapleyVsIG2023, ExplainSHAPorIG2023} highlight trade-offs in fidelity, runtime, and reproducibility. Efforts to improve efficiency through sampling-based approximations or surrogate models, such as FastSHAP \cite{EfficientShapley2024}, introduce new challenges around approximation accuracy and training stability.

Our approach, DeepACTIF, introduces a novel class of attribution grounded in internal network activations. By leveraging the statistical properties of activations within LSTM layers and applying an inverse-weighted aggregation strategy, DeepACTIF avoids the pitfalls of gradients and perturbations. Our method is inherently model-aligned, efficient, and robust, requiring no surrogate models or external training. In contrast to averaging-based aggregation, our inverse strategy promotes stability and down-weights noisy or inconsistent signals.

Through extensive empirical benchmarking and statistical testing, we show that DeepACTIF consistently outperforms traditional and state-of-the-art methods under feature-sparse constraints (top-k subsets). This makes it a practical solution for interpretable, efficient, and sustainable AI in sequential domains.
\section{Methods}
\label{sec:method}

\begin{figure*}[!t]
    \centering
    \includegraphics[width=\textwidth]{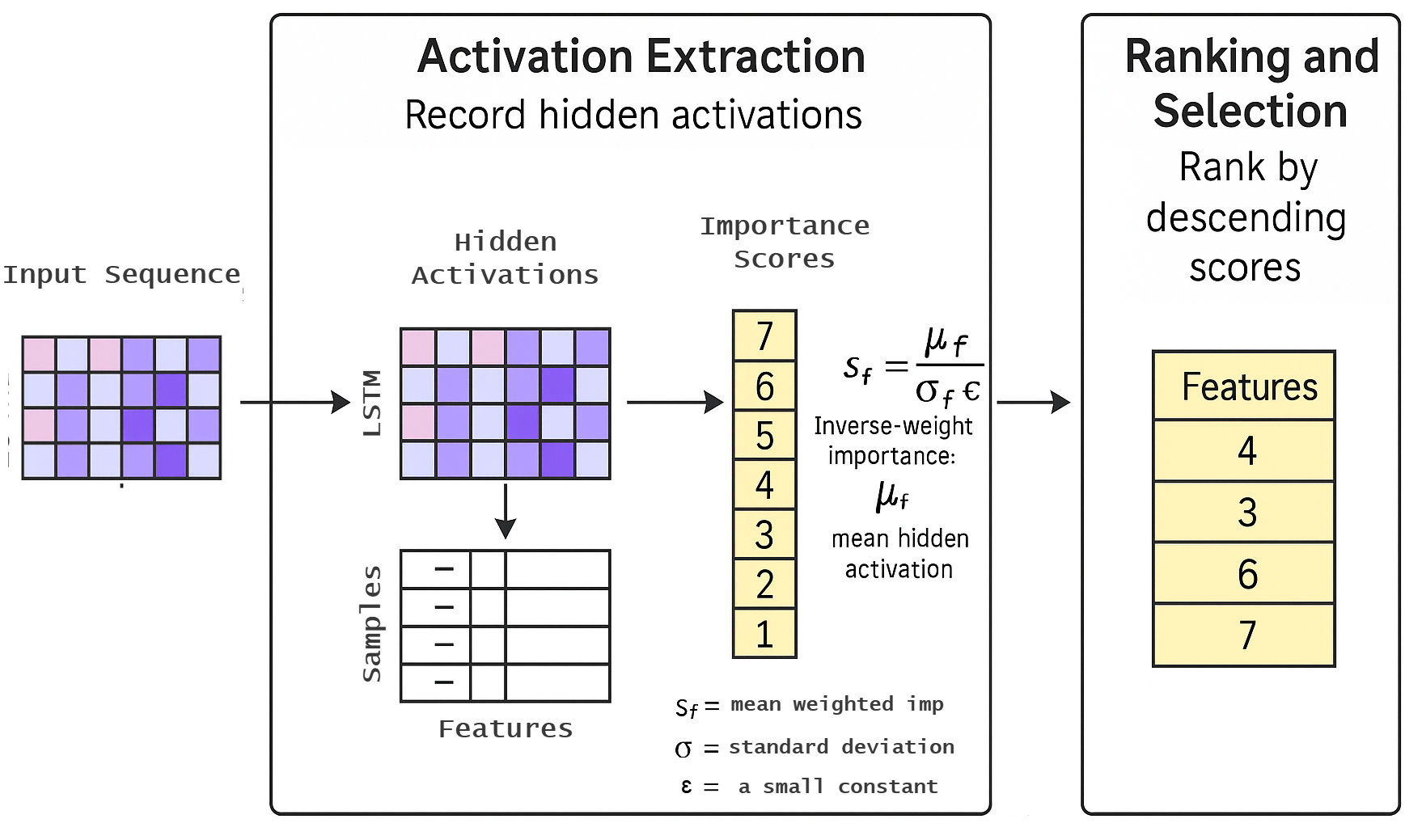}
    \caption{Overview of the DeepACTIF framework. The method extracts hidden activations from a trained LSTM model for each input sequence. It then computes feature importance scores using an inverse-weighted aggregation strategy that favours features with consistently strong activations across samples. Features are ranked by their importance scores and selected for downstream tasks, enabling efficient and interpretable feature attribution without gradients or perturbations.}
    \label{fig:deepactif_framework}
\end{figure*}

DeepACTIF is a lightweight and interpretable feature attribution framework designed explicitly for LSTM-based sequence models. As illustrated in Figure~\ref{fig:deepactif_framework}, DeepACTIF attributes importance scores to input features by leveraging internal activations of the network, without relying on gradients, perturbations, or surrogate models. This makes it particularly well-suited for real-time and low-resource applications.

\subsection{Overview}

Given a trained LSTM model and a dataset of multivariate sequences $\{\mathbf{x}^{(i)}\}_{i=1}^N$, where each sequence $\mathbf{x}^{(i)} \in \mathbb{R}^{T \times F}$ has $T$ timesteps and $F$ features, DeepACTIF proceeds in three steps:
\begin{enumerate}
    \item Capture hidden activations from a designated layer (e.g., LSTM or input).
    \item Aggregate activations over time and across samples using an inverse-weighted strategy.
    \item Rank features by importance and evaluate the model on top-$k\%$ feature subsets.
\end{enumerate}

\subsection{Layer Activation Capture}

For each input sequence, we record the hidden activations $\mathbf{A}^{(i)}$ from a selected layer (typically the LSTM or input layer). These activations reflect the model’s internal representation of the input signal.

Let:
\[
\mathbf{A}^{(i)} \in \mathbb{R}^{T \times H}
\]
denote the activation tensor of sample $i$, with $H$ units. To map these activations back to input features $f \in \{1, \dots, F\}$, we rely on the layer’s input connections or retain the input itself for the input-layer variant.

\subsection{Inverse-Weighted Aggregation (INV Strategy)}

We summarise the relevance of each feature $f$ across all timesteps and sequences using an inverse-weighted mean formulation:
\[
s_f = \frac{\mu_f}{\sigma_f + \epsilon}
\]
where\\
 $\mu_f$ is the mean activation for feature $f$ across all samples and timesteps.\\
 $\sigma_f$ is the standard deviation (measuring instability or noise).\\
 $\epsilon$ is a small constant to avoid division by zero.\\

This formulation prioritises features that are \textbf{consistently active} (high mean, low variance), resulting in a robust global feature ranking.

The inverse-weighted aggregation strategy in DeepACTIF is grounded in the intuition that important features tend to produce consistent, high-magnitude activations across sequences. Features with a high mean activation indicate a strong influence on the network's internal representation, while low variance suggests stability and reliability in their contribution. By combining these two aspects, the strategy emphasises features that are both salient and robust. This aligns with findings in interpretability research, where consistent internal representations are often linked to semantically meaningful or causally relevant inputs. In contrast, noisy or erratically active features, those with high variance but low mean, are downweighted, as they are less likely to reflect stable model reasoning. This mechanism can also be seen as a simplified proxy for reducing activation entropy, encouraging focus on features that reliably drive model behaviour.


After computing $s_f$ scores for all features, we rank them in descending order to obtain a global importance ordering. To evaluate attribution quality, we perform \textbf{top-$k$ feature selection}: for each method, we retrain and evaluate the model using only the top $k\%$ of features ($k \in \{10, 20, 30, 40\}$). Performance degradation (as measured by MAE) serves as a proxy for attribution fidelity.
\begin{algorithm}[H]
\caption{DeepACTIF: Inverse-Weighted Feature Importance}
\label{alg:deepactif}
\begin{algorithmic}[1]
\Require Trained LSTM model $M$, input sequences $\{\mathbf{x}^{(i)} \in \mathbb{R}^{T \times F}\}_{i=1}^N$
\Ensure Feature importance scores $\{s_f\}_{f=1}^F$

\State Initialize importance vector $s \gets \mathbf{0}\in\mathbb{R}^F$

\For{each sequence $\mathbf{x}^{(i)}$}
    \State Extract hidden activations $\mathbf{A}^{(i)} \in \mathbb{R}^{T \times H}$ from target layer
    \State Map activations back to input features
    \State Accumulate per-feature activation statistics over time
\EndFor

\For{$f = 1,\dots,F$}
    \State Compute mean $\mu_f$ and standard deviation $\sigma_f$ across all $T \times N$ values
    \State Compute score $s_f \gets \dfrac{\mu_f}{\sigma_f + \epsilon}$
\EndFor

\State \Return Sorted feature scores $\{s_f\}$ in descending order
\end{algorithmic}
\end{algorithm}

\subsection{Advantages}

Compared to gradient-based and sampling-based attribution methods, DeepACTIF offers the following benefits:
\begin{itemize}
    \item \textbf{Efficiency:} Requires only forward activations; no backpropagation or perturbation.
    \item \textbf{Robustness:} Penalises noisy activations, favouring consistent contributors.
    \item \textbf{Simplicity:} Operates directly on internal activations without auxiliary models.
    \item \textbf{Scalability:} Easily deployable in real-time or edge scenarios due to low overhead.
\end{itemize}

\section{Experimental Setup}
\label{sec:experiments}

To evaluate the performance of DeepACTIF, we conducted extensive experiments on three real-world gaze datasets, each collected in controlled environments and tailored toward depth estimation tasks. These datasets, GIW (Gaze in the Wild) \cite{kothari2020gaze}, RobustVision\cite{hosp2024simulation}, and Tufts\cite{Stone2023Gaze}, contain multivariate time-series sequences of gaze features labelled with continuous depth targets. After applying subject-wise normalisation and segmenting the data into fixed-length sequences, the resulting inputs preserve the temporal structure necessary for sequence modelling.

For all experiments, we employed a single-layer Long Short-Term Memory (LSTM) network followed by a fully connected prediction head. This architecture was chosen for its proven effectiveness in modelling temporal dependencies in gaze behaviour. Models were trained using the AdamW optimiser with a learning rate of \(1 \times 10^{-3}\) and a batch size of 460. Smooth L1 loss was used as the objective, and early stopping was applied based on validation MAE. Hyperparameters were selected using cross-validation.

To ensure a robust and subject-independent evaluation, we adopted a leave-one-out cross-validation (LOOCV) strategy. In each fold, the model was trained on data from all participants except one, which was held out for testing. This approach captures subject variability and ensures generalizability across individuals.

We compared DeepACTIF to several established feature attribution methods, including Integrated Gradients (IG), DeepLIFT, and SHAP. For SHAP, we explored three configurations that trade off between time, memory, and precision. Perturbation-based baselines such as Ablation and Shuffle were also included, along with Random and Zero as control conditions. DeepACTIF itself was evaluated at three levels of the network architecture: the input layer, the LSTM output, and the penultimate fully connected layer.

Each method produced a ranking of feature importance scores, from which we generated feature subsets containing the top \(k\%\) of features, with \(k \in \{10, 20, 30, 40\}\). To assess the fidelity of these rankings, we retrained the model using only the selected features and evaluated the resulting performance. We measured accuracy using Mean Absolute Error (MAE), statistical robustness using Wilcoxon signed-rank tests and Cohen’s \(d\), and efficiency in terms of average execution time and memory consumption. All experiments were repeated across 25 random runs using identical splits and seeds to ensure fair comparisons.

All models and evaluation pipelines were implemented in Python using PyTorch and Scikit-learn. The experiments were run on a system equipped with an Intel Core i7-118900H processor, an NVIDIA RTX 3070 Ti GPU, and 16 GB DDR4 RAM, ensuring sufficient capacity for even the most computationally intensive methods like SHAP.

\section{Results}
\label{sec:results}

We present the experimental results of DeepACTIF across three key aspects: predictive performance using reduced feature sets, statistical significance of performance differences, and computational efficiency in terms of memory and runtime.

\subsection{Top-$k$ Feature Selection Performance}

\begin{figure*}[!ht]
    \centering
    \includegraphics[width=\textwidth]{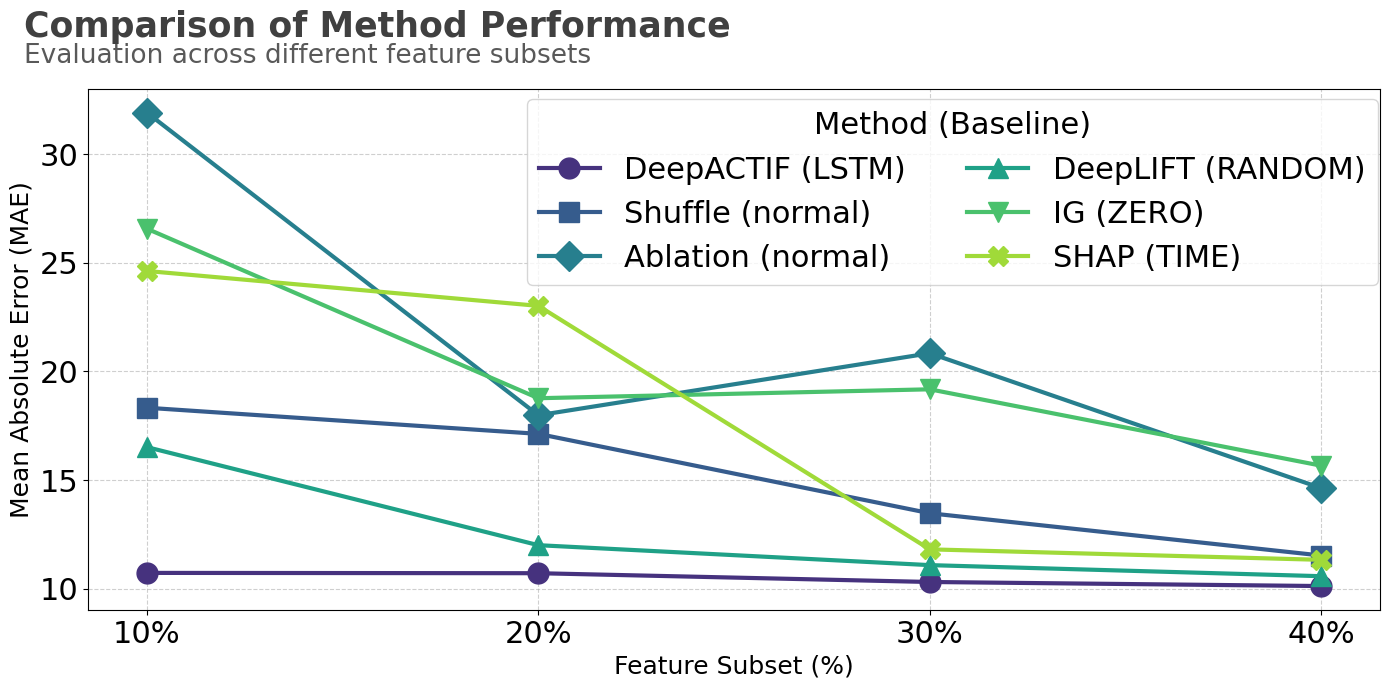}
    \caption{Performance comparison across feature subsets. DeepACTIF (\texttt{LSTM}) consistently outperformed other methods, especially when using only 10–20\% of input features.}
    \label{fig:accuracy_comp}
\end{figure*}

Figure~\ref{fig:accuracy_comp} illustrates the Mean Absolute Error (MAE) for each method as a function of the percentage of top-ranked features retained. Across all feature subset sizes, DeepACTIF, specifically the variant using the inverse-weighted (INV) aggregation strategy at the LSTM layer, consistently outperformed all baseline methods. The performance margin was particularly pronounced in low-resource settings, such as when using only the top 10\% or 20\% of features. These results indicate that DeepACTIF not only identifies the most informative features but also maintains predictive accuracy under strong dimensionality constraints.

We evaluated whether DeepACTIF (LSTM) significantly outperformed competing methods using the Wilcoxon signed-rank test. Table~\ref{tab:statistical_comparison} reports $p$-values and Cohen’s $d$ effect sizes for each comparison across top-$k$ feature subsets. 

Results show that DeepACTIF significantly outperforms Ablation and SHAP in most configurations (e.g., $p<0.001$, $|d|>0.8$), with moderate to large effect sizes. While some variants of DeepLIFT and IG remain competitive at higher $k$ levels, DeepACTIF consistently ranks among the top-performing methods under strict feature constraints.

\begin{table}[ht]
\centering
\small
\begin{tabular}{lllll}
\toprule
\textbf{Method} & \textbf{Top-10\%} & \textbf{Top-20\%} & \textbf{Top-30\%} & \textbf{Top-40\%} \\
\midrule
Ablation (normal)     & p=0.0018, d=-0.76 & p<0.001, d=-0.92  & p<0.001, d=-0.80  & p=0.0012, d=-0.72 \\
DeepLIFT (RANDOM)     & p=0.0056, d=0.62  & p=0.5249, d=-0.07 & p=0.8532, d=-0.02 & p=0.7510, d=0.10  \\
IntGrad (RANDOM)      & p<0.001, d= -    & p<0.001, d= -    & p<0.001, d= -    & p=0.4578, d=-0.06 \\
IntGrad (ZERO)        & p=0.1014, d=-0.36 & p<0.001, d=-1.09  & p<0.001, d=-0.60  & p<0.001, d= -     \\
SHAP (TIME)           & p=0.4108, d=-0.09 & p<0.001, d=-1.32  & p=0.1199, d=-0.23 & p=0.3525, d=-0.13 \\
Shuffle (normal)      & p=0.2996, d=0.28  & p<0.001, d=-0.82  & p=0.0667, d=-0.37 & p=0.6528, d=-0.11 \\
\bottomrule
\end{tabular}
\caption{Wilcoxon signed-rank $p$-values and Cohen’s $d$ effect sizes comparing DeepACTIF (LSTM) to the best variant of each baseline method. Cells marked “—” indicate undefined effect sizes due to zero variance across samples.}

\label{tab:statistical_comparison}
\end{table}
    
\begin{figure}[t]
    \centering
    \includegraphics[width=0.95\linewidth]{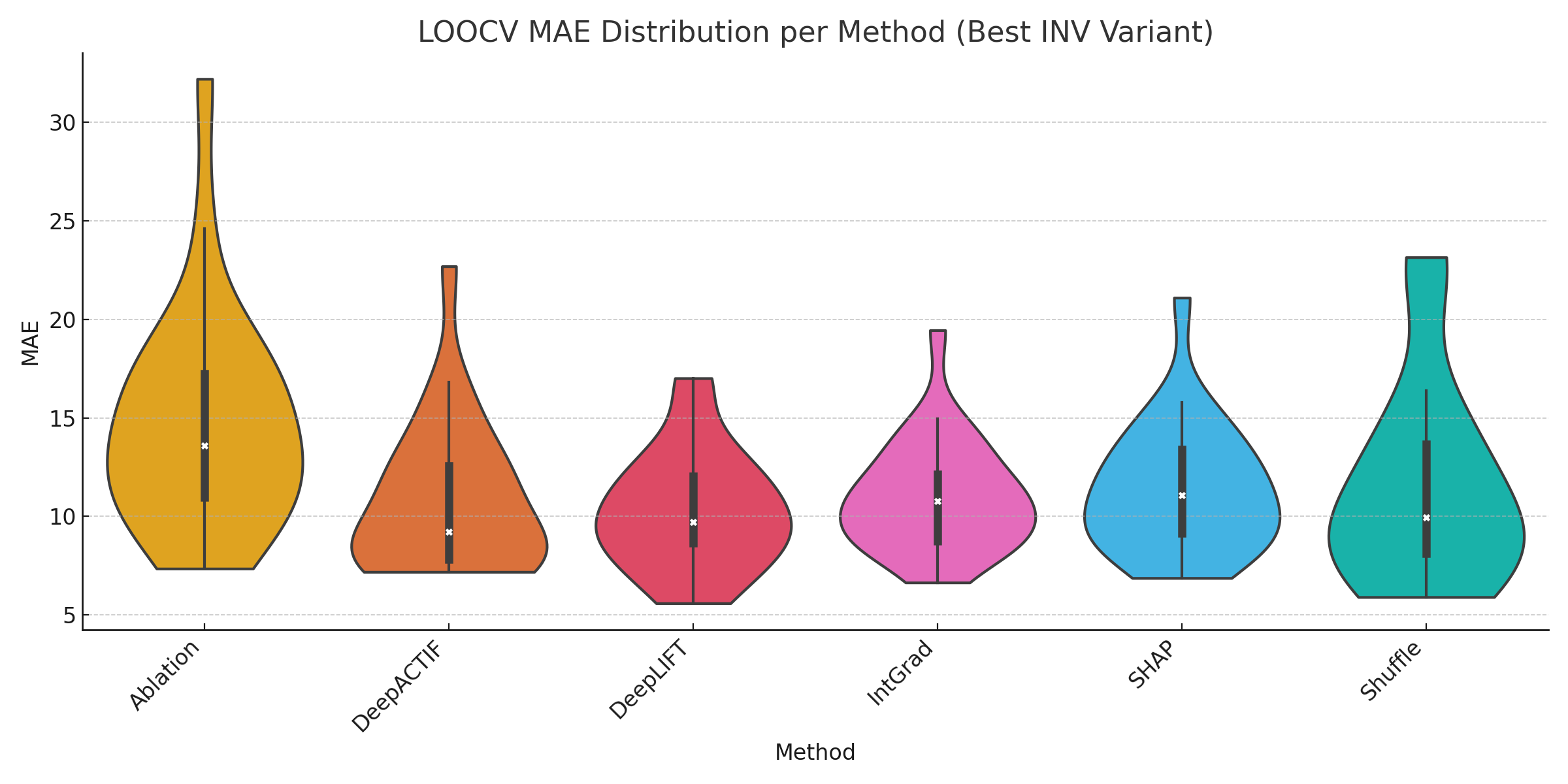}
    \caption{MAE distribution across LOOCV folds for each attribution method. DeepACTIF achieves both low error and consistent results.}
    \label{fig:violin_loocv}
\end{figure}

To visualise the consistency and spread of attribution performance, Figure~\ref{fig:violin_loocv} shows the distribution of MAE across LOOCV folds for each method. DeepACTIF (LSTM) exhibits both low average MAE and narrow variance across subjects, indicating robust feature selection. In contrast, many baseline methods show high variance or outliers, suggesting less stable attribution quality. This visualisation complements the top-$k$ performance analysis by highlighting not only which methods perform well on average, but also how reliably they do so.

\subsection{Computational and Memory Efficiency}

\begin{figure*}[!ht]
    \centering
    \includegraphics[width=\textwidth]{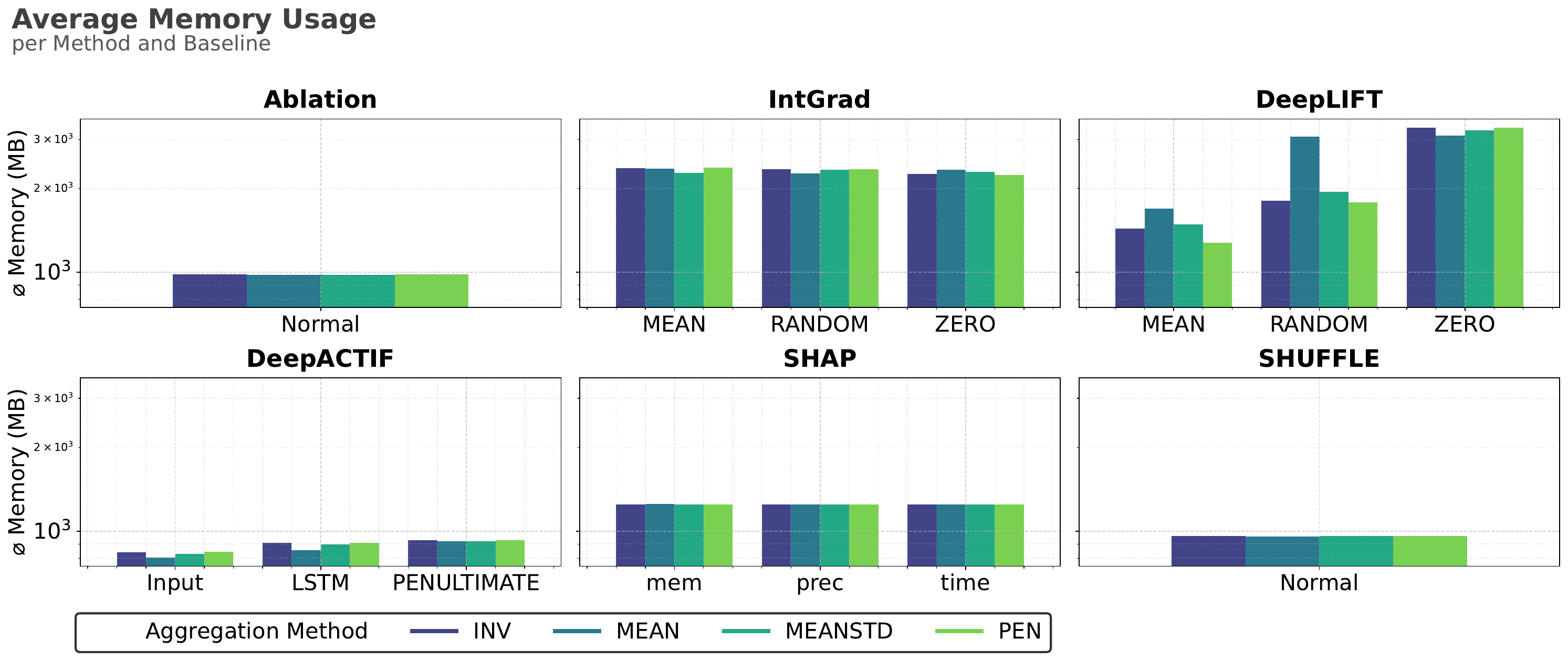}
    \caption{Average memory usage per method. DeepACTIF variants require significantly less memory, enabling use on edge devices.}
    \label{fig:mem_comp}
\end{figure*}

In addition to predictive performance, DeepACTIF offers considerable efficiency advantages. Figure~\ref{fig:mem_comp} shows the average memory usage per method. While methods such as SHAP and IG consume several hundred megabytes per subject due to backpropagation and large sample sets, DeepACTIF variants remain consistently lightweight. 

\begin{figure*}[!ht]
    \centering
    \includegraphics[width=\textwidth]{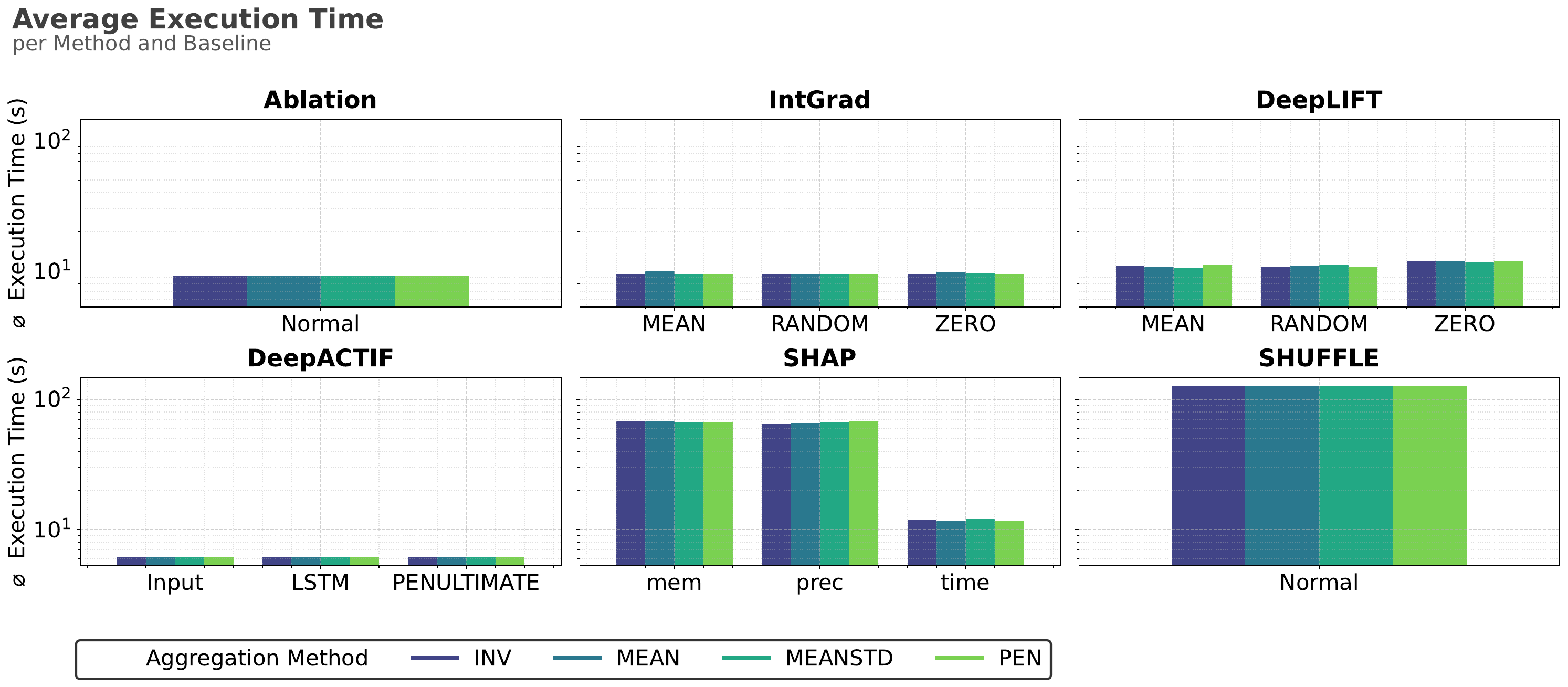}
    \caption{Average execution time per subject. DeepACTIF is significantly faster than gradient- or sampling-based methods, enabling real-time deployment.}
    \label{fig:time_comp}
\end{figure*}

Figure~\ref{fig:time_comp} highlights execution time per method. DeepACTIF completed attribution for a subject in under 6.15 seconds, orders of magnitude faster than SHAP (65.12 seconds in its precision-optimised configuration). These results demonstrate the suitability of DeepACTIF for real-time applications and low-power systems.

DeepACTIF demonstrates substantial advantages in both runtime and memory usage compared to gradient- and sampling-based attribution methods. Across 25 repeated runs, DeepACTIF (LSTM) achieved an average execution time of only 6.15 seconds per subject. It required just 907 MB of memory, making it nearly 2× faster and significantly more memory-efficient than common gradient-based methods. For example, Integrated Gradients (MEAN) required 9.38 s and 2364 MB, and DeepLIFT (ZERO) took 12.02 s while consuming over 3.3 GB of memory. Sampling-based methods like SHAP were even more computationally intensive: SHAP (PRECISION) took 65.12 s per subject, and SHUFFLE required 126.54 s on average. These results highlight DeepACTIF’s suitability for real-time and resource-constrained environments where low latency and minimal memory overhead are critical.

In terms of memory usage, DeepACTIF consistently remained under \textbf{1 GB}, whereas SHAP, IG, and DeepLIFT methods often exceeded \textbf{2–3 GB}, with some variants peaking above 3 GB. These results emphasise the suitability of DeepACTIF for real-time deployment and resource-constrained environments, such as mobile eye tracking or embedded medical systems, where both latency and energy efficiency are critical.
See Appendix \ref{appendix:rawperformance} and \ref{appendix:rawtimeandmemory} for raw performance and runtime statistic
\section{Discussion}
\label{sec:discussion}

Our evaluation demonstrates that \textbf{DeepACTIF} is a robust and efficient feature attribution method tailored to sequence models operating under computational constraints. Across three real-world gaze datasets, DeepACTIF consistently outperformed established methods in identifying impactful features, even when restricted to the top 10–20\% of the input space. These gains were most pronounced when using the inverse-weighted (\texttt{INV}) aggregation strategy applied at the LSTM layer, which effectively captured temporally stable and informative signals.

Beyond its predictive accuracy, DeepACTIF offers substantial computational advantages. By avoiding backpropagation and input perturbation, memory usage and execution time are dramatically reduced compared to gradient- and sampling-based methods like SHAP or Integrated Gradients. These improvements make DeepACTIF especially well-suited for deployment in latency-sensitive and resource-limited environments, such as wearable assistive systems or mobile diagnostic tools.

Crucially, DeepACTIF also advances interpretability. Its \texttt{INV} strategy promotes features that activate reliably across sequences and samples, aligning with human intuitions about stable contributors versus noisy outliers. This results in more robust and trustworthy feature rankings, an essential trait for decision-critical applications. Intuitively, the \texttt{INV} aggregation strategy balances magnitude and stability by dividing each feature’s average activation by its variability. This favours inputs that are both strong and consistent contributors, filtering out those that may exhibit high but erratic influence. Such regularity is a valuable trait in time-series settings, where transient spikes may be misleading, and consistent activations better reflect underlying model reasoning.

Nevertheless, this work has several limitations that open avenues for future research. While DeepACTIF is designed to be architecture-agnostic, our experiments focused exclusively on LSTM-based models in gaze-based regression tasks. Although the methodology is readily applicable to other architectures—such as Transformers or CNNs—and other domains beyond gaze, these extensions remain to be validated empirically. Furthermore, our evaluation framework emphasises fidelity by measuring model performance on top-ranked features, but does not yet incorporate causal or counterfactual tests to assess attribution faithfulness. Evaluating DeepACTIF under distributional shifts, adversarial settings, or in human-in-the-loop scenarios would strengthen its reliability. Finally, while the three datasets used in this study are diverse, they all originate from biometric tracking contexts, which may limit generalizability to unrelated domains.

Overall, DeepACTIF provides a practical and principled approach to interpretable feature attribution in sequential data. Future work may explore its extension to classification tasks, multi-modal learning (e.g., gaze and video), and real-world applications such as longitudinal health monitoring, thereby broadening its utility and impact.

\section{Conclusion}
\label{sec:conclusion}

We present \textbf{DeepACTIF}, a lightweight, interpretable, and resource-efficient framework for feature attribution in sequential deep learning models. By leveraging internal activations and introducing a novel inverse-weighted aggregation strategy, DeepACTIF achieves state-of-the-art results across predictive accuracy, computational speed, and memory usage. Its design supports deployment in real-time, low-latency environments where traditional methods fall short.

Although our current study focuses on LSTM models, the principles behind DeepACTIF are readily extensible. The method requires only access to hidden activations, enabling straightforward adaptation to other architectures such as Transformers, CNNs, and MLPs. Preliminary results suggest feasibility in these directions, and future work will validate these extensions and explore cross-domain applicability.

DeepACTIF provides a practical solution to a longstanding trade-off between interpretability and efficiency. Enabling reliable attribution with minimal computational cost contributes meaningfully to the broader goal of trustworthy and accessible AI, especially in domains where transparency, responsiveness, and resource awareness are all critical.

\subsection{Code and Dataset Availability}

To promote transparency and encourage community adoption, all code and datasets used in this study are publicly available at \url{https://github.com/benedikt-hosp/actif}. Reproducibility and open collaboration remain at the core of this work, fostering further innovation in interpretable AI.

\bibliography{actif}

\begin{thebibliography}{10}

\bibitem{ShapleyVsIG2023}
Tianshu Feng, Zhipu Zhou, Joshi Tarun, and Vijayan~N Nair.
\newblock Comparing baseline shapley and integrated gradients for local explanation: Some additional insights.
\newblock 2022.

\bibitem{ExplainSHAPorIG2023}
Frank Frankowski and Taly Ankur.
\newblock Should you explain your predictions with shap or ig?
\newblock {\em Fiddler AI Blog}, 2023.
\newblock Available online: \url{https://www.fiddler.ai}.

\bibitem{hosp2024simulation}
Benedikt~W Hosp, Yannick Sauer, Bj{\"o}rn Severitt, Rajat Agarwala, and Siegfried Wahl.
\newblock Simulation of various tuning methods in autofocals using a virtual reality headset.
\newblock {\em Optics Continuum}, 3(8):1273--1290, 2024.

\bibitem{kothari2020gaze}
Rakshit Kothari, Zhizhuo Yang, Christopher Kanan, Reynold Bailey, Jeff~B Pelz, and Gabriel~J Diaz.
\newblock Gaze-in-wild: A dataset for studying eye and head coordination in everyday activities.
\newblock {\em Scientific reports}, 10(1):2539, 2020.

\bibitem{EfficientShapley2024}
Xiaoxiao Li, Yuan Zhou, Nicha~C Dvornek, Yufeng Gu, Pamela Ventola, and James~S Duncan.
\newblock Efficient shapley explanation for features importance estimation under uncertainty.
\newblock In {\em Medical Image Computing and Computer Assisted Intervention--MICCAI 2020: 23rd International Conference, Lima, Peru, October 4--8, 2020, Proceedings, Part I 23}, pages 792--801. Springer, 2020.

\bibitem{lundberg2017unified}
Scott~M Lundberg and Su-In Lee.
\newblock A unified approach to interpreting model predictions.
\newblock {\em Advances in neural information processing systems}, 30, 2017.

\bibitem{Meyes2019Ablation}
R.~Meyes, Melanie Lu, C.~W.~D. Puiseau, and Tobias Meisen.
\newblock Ablation studies in artificial neural networks.
\newblock {\em ArXiv}, abs/1901.08644, 2019.

\bibitem{Radivojac2004Feature}
P.~Radivojac, Z.~Obradovic, A.~Dunker, and S.~Vucetic.
\newblock Feature selection filters based on the permutation test.
\newblock pages 334--346, 2004.

\bibitem{Shawi2020Interpretability}
Radwa~El Shawi, Youssef Mohamed, M.~Al-Mallah, and S.~Sakr.
\newblock Interpretability in healthcare: A comparative study of local machine learning interpretability techniques.
\newblock {\em Computational Intelligence}, 37:1633 -- 1650, 2020.

\bibitem{shrikumar2017learning}
Avanti Shrikumar, Peyton Greenside, and Anshul Kundaje.
\newblock Learning important features through propagating activation differences.
\newblock In {\em International conference on machine learning}, pages 3145--3153. PMlR, 2017.

\bibitem{Stone2023Gaze}
Abigail Stone, Srijith Rajeev, S.~P. Rao, K.~Panetta, S.~Agaian, Aaron Gardony, Jessica Nordlund, and Rebecca Skantar.
\newblock Gaze depth estimation for eye-tracking systems.
\newblock 12526:125260N--125260N--10, 2023.

\bibitem{sundararajan2017axiomatic}
Mukund Sundararajan, Ankur Taly, and Qiqi Yan.
\newblock Axiomatic attribution for deep networks.
\newblock In {\em International conference on machine learning}, pages 3319--3328. PMLR, 2017.

\bibitem{Tjoa2019A}
Erico Tjoa and Cuntai Guan.
\newblock A survey on explainable artificial intelligence (xai): Toward medical xai.
\newblock {\em IEEE Transactions on Neural Networks and Learning Systems}, 32:4793--4813, 2019.

\bibitem{EvaluationTimeSeries2023}
Hugues Turb{\'e}, Mina Bjelogrlic, Christian Lovis, and Gianmarco Mengaldo.
\newblock Evaluation of post-hoc interpretability methods in time-series classification.
\newblock {\em Nature Machine Intelligence}, 5(3):250--260, 2023.

\bibitem{SurveyTimeSeries2023}
Ziqi Zhao, Yucheng Shi, Shushan Wu, Fan Yang, Wenzhan Song, and Ninghao Liu.
\newblock Interpretation of time-series deep models: A survey.
\newblock {\em arXiv preprint arXiv:2305.14582}, 2023.

\end{thebibliography}
\bibliographystyle{plain}

\appendix

\clearpage

\appendix
\section{Appendix}

\subsection{Raw Performance Data}
\label{appendix:rawperformance}
\renewcommand{\arraystretch}{1.1}
\begin{longtable}{lllcccc}

\toprule
\textbf{Method} & \textbf{Baseline}  & \textbf{10\% MAE} & \textbf{20\% MAE} & \textbf{30\% MAE} & \textbf{40\% MAE} \\
\midrule
\endfirsthead

\toprule
\textbf{Method} & \textbf{Baseline} & \textbf{10\% MAE} & \textbf{20\% MAE} & \textbf{30\% MAE} & \textbf{40\% MAE} \\
\midrule
\endhead

\bottomrule
\multicolumn{7}{r}{\textit{Continued on the next page...}} \\
\endfoot

\bottomrule
\endlastfoot

\multirow[t]{4}{*}{Ablation} & \multirow[t]{4}{*}{-} &  31.90 & 17.98 & 20.83 & 14.63 \\

\cline{1-7} \cline{2-7}
\multirow[t]{12}{*}{DeepACTIF} & \multirow[t]{4}{*}{Input} & 23.79 & 19.86 & 12.71 & 11.37 \\

\cline{2-7}
 & \multirow[t]{4}{*}{LSTM} & 10.73 & 10.71 & 10.31 & 10.12 \\

\cline{2-7}
 & \multirow[t]{4}{*}{Penultimate} &  19.30 & 18.60 & 19.92 & 11.71 \\
\cline{1-7} \cline{2-7}
\multirow[t]{12}{*}{DeepLIFT} & \multirow[t]{4}{*}{MEAN} & 29.66 & 23.54 & 22.02 & 21.09 \\
\cline{2-7}
 & \multirow[t]{4}{*}{RANDOM} &  16.51 & 12.00 & 11.09 & 10.58 \\
\cline{2-7}
 & \multirow[t]{4}{*}{ZERO} &  34.73 & 19.96 & 16.83 & 11.28 \\
\cline{1-7} \cline{2-7}
\multirow[t]{12}{*}{IG} & \multirow[t]{4}{*}{MEAN} & 29.53 & 29.34 & 25.54 & 18.70 \\
\cline{2-7}
 & \multirow[t]{4}{*}{RANDOM} &  37.16 & 25.10 & 22.77 & 11.00 \\
\cline{2-7}
 & \multirow[t]{4}{*}{ZERO} & 26.56 & 18.76 & 19.17 & 15.66 \\
\cline{1-7} \cline{2-7}
\multirow[t]{12}{*}{SHAP} & \multirow[t]{4}{*}{MEM} & 29.96 & 23.52 & 21.41 & 16.26 \\
\cline{2-7}
 & \multirow[t]{4}{*}{PREC} &  39.32 & 37.43 & 12.11 & 11.51 \\
\cline{2-7}
 & \multirow[t]{4}{*}{TIME} &  24.62 & 23.01 & 11.81 & 11.33 \\
\cline{1-7} \cline{2-7}
\multirow[t]{4}{*}{Shuffle} & \multirow[t]{4}{*}{-} & 18.33 & 17.12 & 13.47 & 11.53 \\
 \caption{Raw MAE results across feature attribution methods, baselines, and aggregation strategies. Values reflect average model performance when using only the top-\(k\%\) ranked features.}
\label{tab:raw_mae}
\end{longtable}

\subsection{Raw Timing and Memory Data}
\label{appendix:rawtimeandmemory}
\begin{longtable}{lllcc}
\toprule
\textbf{Method} & \textbf{Baseline} & \textbf{Avg. Exec. Time} & \textbf{Avg. Memory Usage} \\
\midrule
\endfirsthead

\toprule
\textbf{Method} & \textbf{Baseline} &  \textbf{Avg. Exec. Time} & \textbf{Avg. Memory Usage} \\
\midrule
\endhead

\bottomrule
\multicolumn{5}{r}{\textit{Continued on the next page...}} \\
\endfoot

\bottomrule
\endlastfoot

Ablation & Normal &  9.23 & 982.89 \\
\midrule
IG & ZERO &  9.46 & 2258.62 \\
   & RANDOM &  9.46 & 2341.97 \\
   & MEAN &  9.38 & 2364.28 \\
\midrule
DeepLIFT & MEAN &  10.91 & 1438.18 \\
         & RANDOM &  10.72 & 1806.63 \\
         & ZERO &  12.02 & 3306.42 \\
\midrule
DeepACTIF & Input &  6.13 & 839.18 \\
          & LSTM &  6.15 & 907.35 \\
          & PENULTIMATE & 6.16 & 927.09 \\
\midrule
SHAP & mem &  68.50 & 1249.01 \\
     & prec &  65.12 & 1248.99 \\
     & time &  11.99 & 1249.14 \\
\midrule
SHUFFLE & Normal  & 126.54 & 960.78 \\
\caption{Execution time and peak memory consumption per method and strategy. All values are averaged over 25 folds.}
\label{tab:time_memory}
\end{longtable}


\newpage

\end{document}